\documentclass[conference]{IEEEtran}
\IEEEoverridecommandlockouts
\usepackage{etoolbox}
\makeatletter
\patchcmd{\@makecaption}
  {\\}
  {.\ }
  {}
  {}
\makeatletter
\patchcmd{\@makecaption}
  {\scshape}
  {}
  {}
  {}
\makeatother
\def\tablename{Table}
\usepackage{amsmath,amssymb,amsfonts}
\usepackage{algorithmic}
\usepackage{graphicx}
\usepackage{biblatex}
\usepackage{textcomp}
\usepackage{adjustbox}
\usepackage{mathrsfs}
\usepackage{bbm}
\usepackage{comment}
\usepackage{xcolor}
\usepackage{mathtools}

\def\BibTeX{{\rm B\kern-.05em{\sc i\kern-.025em b}\kern-.08em
    T\kern-.1667em\lower.7ex\hbox{E}\kern-.125emX}}

\usepackage{fancyhdr}
\begin{document}

\title{Robust Class-Conditional Distribution Alignment for Partial Domain Adaptation}

\author{\IEEEauthorblockN{Sandipan Choudhuri, Arunabha Sen}
\text{Arizona State University}\\
\{s.choudhuri, asen\}@asu.edu}

\maketitle

\begin{abstract}
Unwanted samples from private source categories in the learning objective of a partial domain adaptation setup can lead to negative transfer and reduce classification performance. Existing methods, such as re-weighting or aggregating target predictions, are vulnerable to this issue, especially during initial training stages, and do not adequately address class-level feature alignment. Our proposed approach seeks to overcome these limitations by delving deeper than just the first-order moments to derive distinct and compact categorical distributions. We employ objectives that optimize the intra and inter-class distributions in a domain-invariant fashion and design a robust pseudo-labeling for efficient target supervision. Our approach incorporates a complement entropy objective module to reduce classification uncertainty and flatten incorrect category predictions. The experimental findings and ablation analysis of the proposed modules demonstrate the superior performance of our proposed model compared to benchmarks. 
\end{abstract}

\section{\textbf{Introduction}}

\noindent
Deep neural networks have remarkably enhanced the performance of current machine learning frameworks \cite{liu2021review, wang2019development, dang2019deep, choudhuri2018object, guo2021survey}. However, their generalizability rests on access to large annotated datasets, which are often challenging to obtain. \textit{Domain adaptation (da)} approaches \cite{li2020deep,ganin2016domain} present a solution, allowing for the transfer of knowledge from labeled to unlabeled datasets. Still, a majority of \textit{da} setups \cite{li2020deep,ganin2016domain,ganin2015unsupervised} presuppose identical label space across both domains—a challenging prerequisite in real-world scenarios. \textit{Partial domain adaptation (pda)} \cite{cao2018partial} offers a more versatile approach, accommodating cases where the label set of the source encompasses that of the target.

Within the $pda$ context, a pivotal challenge arises from the absence of label overlap information between the domains. This can inadvertently introduce \textit{negative transfer} \cite{cao2018partial,cao2019learning}, where irrelevant data from the source hampers the target classification. Although conventional strategies, such as re-weighting or aggregating target predictions, have been deployed, they remain vulnerable to errors and noise, especially during the initial stages of training \cite{cao2018partial,zhang2018importance,cao2018partial2,cao2019learning,choudhuri2020partial}. Our proposition counters this by focusing beyond first-order moments \cite{choudhuri2022coupling,choudhuri2023distribution} to align the categorical distributions across domains in a domain-agnostic setup.

A common pitfall in standard domain adaptation is the inadvertent sacrifice of feature discriminability for enhancing feature transferability. This can produce classifiers that, while adept at reducing domain disparities, falter in actual target data classification. Despite the prevalence of standard cross-entropy loss in existing approaches \cite{zhang2018importance, ganin2016domain, cao2018partial2}, some have ventured to address this issue \cite{miyato2018virtual,kumar2018co,shu2018dirt}. These, however, tend to elevate the model's complexity, complicating the training process. In response, our approach integrates a complement entropy objective, ensuring that incorrect classifications are evenly distributed, reducing the likelihood of incorrect categories challenging the ground-truth class.


Additionally, our method utilizes pseudo-labeling to achieve domain and class-level alignment on cross-domain data. The pseudo-labels are generated using a non-trainable prototype classifier to estimate the probability of a sample aligning with a source cluster. Recognizing that initial pseudo-labels might be inconsistent and stray from our goals, a subset of confident target samples, aggregated over a fixed number of iterations that exceed a dynamic classification probability threshold, is subsequently selected for classifier training. This approach ensures high-quality pseudo-labels without increasing the model's trainable parameters.

\begin{figure*}[htbp!]
    \centering
    \includegraphics[width = 0.75\linewidth]{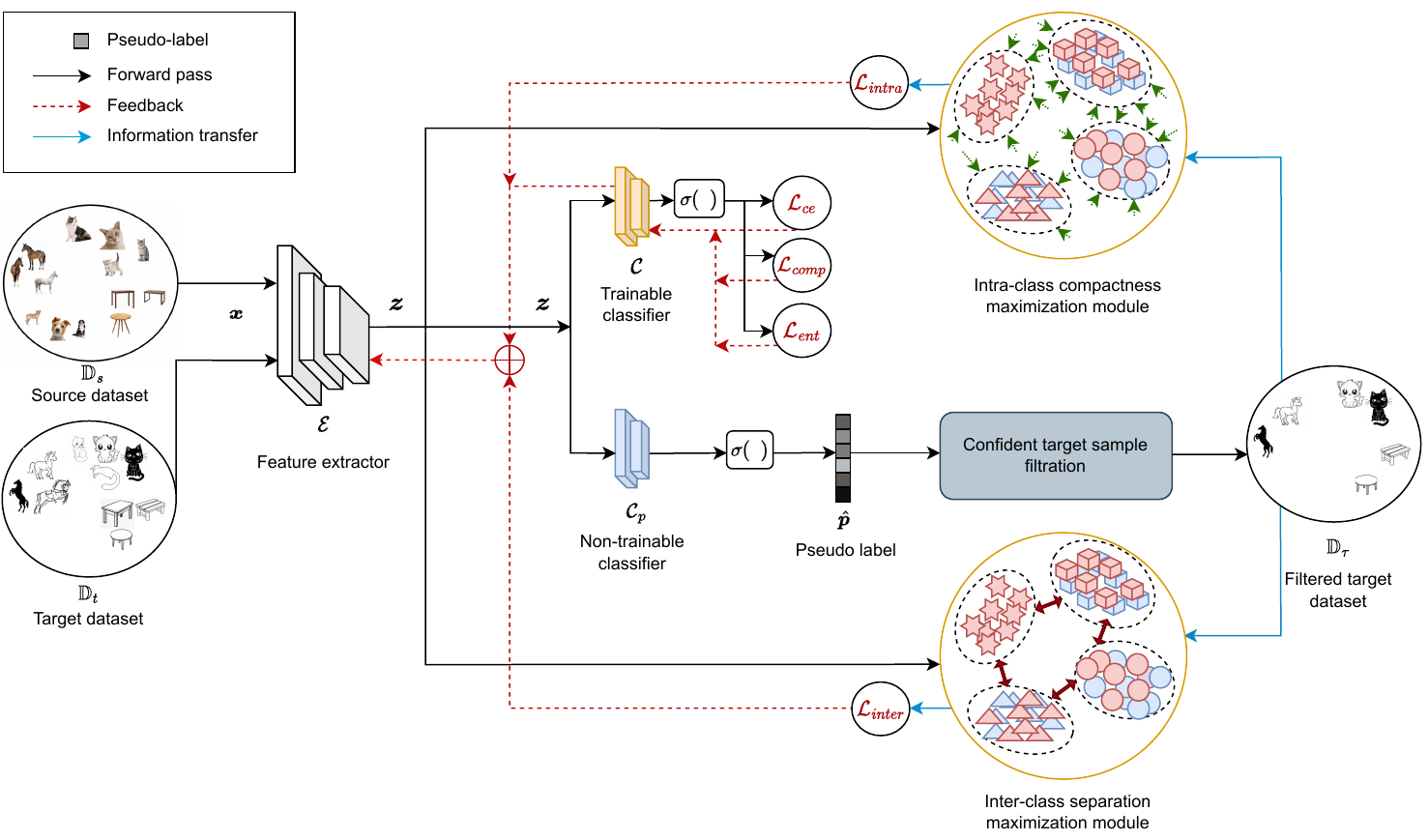}
    \vspace{-2mm}
    \caption{Architectural diagram of the proposed domain adaptation model (model training phase).}    
    \label{proposed_model}  
    \vspace{-4mm}
\end{figure*}

\section{\textbf{Related Works}}

\noindent

\noindent
Numerous studies have addressed domain adaptation to minimize domain discrepancies using labeled data \cite{pan2010survey}. Recent works have utilized deep learning to obtain intricate, transferable features by integrating adversarial loss with domain-invariant data transformation \cite{ganin2016domain, li2019joint}. However, these networks are hard to train, hyper-parameter sensitive, and often restricted to scenarios with identical source and target labels, limiting their utility in a $pda$ context. Relaxing the identical label space assumption introduces the issue of \textit{negative transfer}, a challenge prior models aren't equipped to handle. 

Among the latest state-of-the-art $pda$ frameworks, selective adversarial network models employ multiple adversarial networks to diminish the influence of unique source category samples, enhancing knowledge transfer from categories common between domains \cite{cao2018partial,cao2018partial2}. Subsequent advancements have introduced frameworks for determining class importance and evaluating the transferability of source samples \cite{cao2018partial,cao2018partial2,zhang2018importance,cao2019learning}. These provide a refined metric to differentiate shared from private source categories. However, these models can be vulnerable during initial training phases due to their sensitivity to incorrect model feedback through aggregated noisy predictions, which can hinder classification performance. Some solutions \cite{choudhuri2022coupling,choudhuri2023distribution} aim to align data distributions using distribution means. However, they overlook distribution variability and primarily capture first-order moment insights for cross-domain category distribution alignment. We posit that these methods miss the critical facets of data alignment. Our proposed methodology seeks to rectify these oversights.
 

\section{\textbf{Methodology}}

\subsection{\textbf{Problem Settings}}
\label{problem_settings}

\noindent
This work explores an unsupervised \textit{partial domain adaptation (pda)} scenario. Specifically, our study centers around a labeled source domain, denoted as $S$, and an unlabeled target domain, represented by $T$. The scenario is restricted to a homogeneous setting, implying that the domains share an identical feature space, $\mathcal{X} \subset R^{d_x}$. Considering a discrete source label space $ \mathcal{Y}_s =\{l_k\}_{k=1}^{K_s}$, $ S $ and $ T $ are characterized by the joint distribution $ P(X_s,Y_s) $ and the marginal distribution $ P(X_t) $, respectively (random variables $X_s, X_t \in \mathcal{X}$, and $ Y_s \in \mathcal{Y}_s$). The source and target domains are represented by datasets $\mathbb{D}_s = \big\{\big(\textbf{\textit{x}}^i_s,y^i_s\big)\big\}_{i=1}^{n_s}$ and $\mathbb{D}_t = \big\{\textbf{\textit{x}}^j_t\big\}_{j=1}^{n_t}$, respectively, sampled in an i.i.d. manner from their respective distributions $P(X_s,Y_s)$ and $P(X_t)$. The crux of \textit{pda} is its pertinence to real-world adaptation scenarios wherein there exists a distribution discrepancy between the two domains, and the label space of $S$ subsumes that of $T$ (i.e., $\mathcal{Y}_t \subset \mathcal{Y}_s$). 

Given a multi-class classification task with a hypothesis space $\mathcal{H}$ of scoring functions and a symmetric loss function $\ell: \mathbb{R}^C \times \mathbb{R}^C \rightarrow \mathbb{R}_+$, the objective is the reduce the target classification risk of a hypothesis $h: \mathcal{X}_t \rightarrow \mathcal{Y}_s$ ($h \in \mathcal{H}$), w.r.t. $\ell$, under $P(X_t,Y_t)$. It should be highlighted that while the random variable $Y_t \in \mathcal{Y}_t$, which represents the target label, is utilized for evaluation, it remains unavailable during the adaptation phase. Additionally, utilizing data from the source domain can lead to the \textit{negative-transfer} \cite{cao2018partial2} issue. This problem arises when \textit{samples unique to the source domain, denoted as $\left\{ \left( \textbf{\textit{x}}^i_s,y^i_s \right) \in \mathbb{D}_s \, \big| \, y^i_s \in \mathcal{Y}_s \setminus \mathcal{Y}_t \right\}_{i=1}^{n_s}$, inadvertently transfer irrelevant knowledge, potentially misguiding the classification process.} To mitigate this, it's imperative to judiciously identify categories shared between both domains, aiming to optimize model performance on $\mathbb{D}_t$.

\subsection{\textbf{Proposed Approach}}
\label{proposed_approach}

\noindent
In this work, we aim to conceptualize the classifier hypothesis $h: \mathcal{X} \rightarrow \mathcal{Y}_s$, as the integration of two neural networks: the feature encoder $\mathcal{E}:\mathcal{X} \rightarrow \mathcal{Z}$ transforming the input samples in $\mathcal{X}$ to the latent space $\mathcal{Z} \subset \mathbb{R}^{d_z}$, and the classifier network $\mathcal{C}:\mathcal{Z} \rightarrow \mathbb{R}^{K_s}$ which converts a latent representation $\boldsymbol{z} \in \mathcal{Z}$ into $K_s$ logits. These logits are subsequently processed through a \textit{softmax} ($\sigma$) layer to yield a $K_s$-dimensional probability vector $\boldsymbol{p}$. As shown in eq. \ref{class}, the classification objective is realized using categorical cross-entropy loss $\ell_{ce}(\hspace{0.5mm}\cdot,\cdot\hspace{0.5mm})$, which compares the model's prediction of source samples to the one-hot encoded representation $\boldsymbol{y}_s$ of the respective label $y_s$. Target supervision is realized by employing soft pseudo-labels, denoted as $\hat{\boldsymbol{p}}_t$, derived from a non-parametric prototype classifier $\mathcal{C}_p: \mathcal{Z}\times \mathcal{Z} \rightarrow \mathbb{R}^{K_s}$ (detailed further in sec. \ref{target_supervision}). These soft labels enhance the classification accuracy of $\mathcal{C}(\mathcal{E}(\hspace{0.5mm}\cdot\hspace{0.5mm}))$ over a strategically curated subset $\mathbb{D}_{\tau} \subseteq \mathbb{D}_t$ of $n_{\tau}$ target samples with high-confidence category predictions. The overall classification objective $\mathscr{L}_{ce}$ is represented as follows:

\vspace{-3mm}
\begin{equation}
\begin{gathered}
    \mathscr{L}_{ce}(\theta_{\mathcal{C}},\theta_{\mathcal{E}}) =  
    \frac{1}{n_s}  \sum\limits_{i=1}^{n_s} {\ell_{ce}}(\boldsymbol{p}^i_s, \boldsymbol{y}^i_s) \! + \!\!
     \underset{[\mathbb{D}_{\tau} \neq \varnothing]}{\mathbbm{1}} \frac{1}{n_{\tau}}  \sum\limits_{j=1}^{n_{\tau}} { \ell_{ce}}(\boldsymbol{p}^j_{\tau}, \hat{\boldsymbol{p}}^j_{\tau})\\
    \boldsymbol{p}^i_{s/\tau} \leftarrow \sigma(\mathcal{C}(\mathcal{E}(\boldsymbol{x}^i_{s/\tau})))
\end{gathered}
\label{class}
\end{equation}

\subsubsection{\textbf{Classifier Uncertainty Reduction}}

Cross-entropy has become the go-to training objective for classification in adaptation tasks over time \cite{zhang2018importance, ganin2016domain, cao2018partial2}. It mainly capitalizes on the ground-truth class, sidelining information from incorrect (complement) categories. This neglect doesn't optimize for inter-class separation, leading to uncertainty in classification. For example, in a three-class problem, an output like [0.5, 0.4, 0.1] is more uncertain than [0.5, 0.25, 0.25], even with the same cross-entropy loss, highlighting potential issues near decision boundaries and resulting in incorrect class probabilities that are significant enough to challenge the ground-truth class.


We propose using complement class information to balance predicted probabilities based on recent research on complement objective training \cite{chen2019complement,liang2020balanced}. By averaging entropies of complement classes within a mini-batch, we aim for uniform and low-prediction probabilities. The sample-wise entropy is conditioned on the summation of the predicted probabilities of these incorrect categories. Since our goal is to level out the predictions for $K_s -1$ classes, we aim to maximize their entropy, simplified as minimizing the loss in eq. \ref{comp}. To diminish uncertainty, uncertain samples with higher confidence are prioritized using the $(1-\hat{\boldsymbol{y}}_g)^\gamma$ term, where $\gamma$ regulates emphasis. $k$ represents all classes, excluding the ground truth $g$. For optimizing training, we normalize $\mathscr{L}_{comp}(\theta_{\mathcal{C}},\theta_{\mathcal{E}})$ by the number of complement categories (i.e., $K_s -1$).


\vspace{-3mm}
\begin{equation}
\begin{split}
    \mathscr{L}_{comp}(\theta_{\mathcal{C}},\theta_{\mathcal{E}}) =  
    \frac{1}{K_s-1} \Bigg[\frac{1}{n_s} & \sum\limits_{i=1}^{n_s} { \ell_{comp}}\Big(\boldsymbol{p}^i_s, \boldsymbol{y}^i_s\Big) + \\
    & \underset{[\mathbb{D}_{\tau} \neq \varnothing]}{\mathbbm{1}} \frac{1}{n_{\tau}}  \sum\limits_{j=1}^{n_{\tau}} { \ell_{comp}}\Big(\boldsymbol{p}^j_{\tau}, \hat{\boldsymbol{p}}^j_{\tau}\Big)\Bigg]\\
    \ell_{comp}(\hat{\boldsymbol{y}},\boldsymbol{y}) =  (1-\hat{\boldsymbol{y}}_g)^\gamma & \sum_{k \neq g} \frac{\hat{\boldsymbol{y}}_k}{1-\hat{\boldsymbol{y}}_g} log \frac{\hat{\boldsymbol{y}}_k}{1-\hat{\boldsymbol{y}}_g}
\end{split}
\label{comp}
\end{equation}

\subsubsection{\textbf{Robust Pseudo-label-Based Target Supervision}}
\label{target_supervision}

Aligning class-conditional features across domains while minimizing the negative impact of private source ($\mathcal{Y}_s \setminus \mathcal{Y}_t$) category samples is key to addressing domain distribution discrepancy and negative transfer. To achieve this, we employ a pseudo-labeling-based target supervision approach- building on the advancements in pseudo-labeling \cite{liang2019distant,jing2020adaptively, liang2020we, wu2023unsupervised}, we introduce a non-trainable nearest-centroid classifier, $\mathcal{C}_p$, using cosine similarity of latent features with class centroids (prototypes) and a softmax operation. The source prototypes, $\boldsymbol{\mu} = \big[\boldsymbol{\mu_k}\big]_{k=1}^{K_s}$, are derived from samples $\textbf{\textit{x}}_s \in \mathbb{D}_s$ and updated via an exponential moving average strategy, as given below:


    \begin{equation}
    \begin{gathered}
    \label{mean-calc}
        \boldsymbol{\mu}_{k}^{update} \leftarrow \frac{\sum_{i=1}^{n_{s}} \mathbbm{1}_{y^i_{s} = l_k}\mathcal{E}(\textbf{\textit{x}}^i_{s})}{\sum_{i=1}^{n_{s}} \mathbbm{1}_{y^i_{s} = l_k}}\\
        \boldsymbol{\mu}_{k} \leftarrow \omega \boldsymbol{\mu}_{k}^{update} +(1-\omega)\boldsymbol{\mu}_k
    \end{gathered}
    \end{equation}

\vspace{1mm}
\noindent
Drawing from the efficacy of confidence-guided self-training \cite{zou2019confidence}, we adopt a similar approach to derive soft pseudo-labels $\hat{\boldsymbol{p}}_t$ for samples $\textbf{\textit{x}}_t \in \mathbb{D}_t$, referenced in the objectives of eq. \ref{class}, \ref{comp}. This approach minimizes the adverse effects of noisy one-hot pseudo-labels, especially during initial training phases.    


\vspace{-3mm}
\begin{equation}
\begin{gathered}
    \hat{\boldsymbol{y}}^j_t \leftarrow one\text{-}hot(\hat{\boldsymbol{p}}^j_t)\\
    \hat{\boldsymbol{p}}^j_t \leftarrow  \sigma\big(\mathcal{C}_p(\boldsymbol{x}^j_t,\boldsymbol{\mu})\big) = \sigma\Big( \big[cos\big(\mathcal{E}(\textbf{\textit{x}}^j_t), \boldsymbol{\mu}_{k}\big)\big]_{k=1}^{K_s}\Big)
\end{gathered}
\end{equation}

\vspace{1mm}
\noindent
In the initial learning phase, the existing discrepancy between source and target distributions often results in noisy pseudo-labels, hampering classification accuracy. The classifier prediction confidence $max(\hat{\boldsymbol{p}}_t)$ gauges the quality of a category assignment, with low scores suggesting model confusion. We utilize it to probe a target sample's likelihood of being mapped to its closest cluster center, limiting target supervision to highly confident samples. Leveraging a $K_s$-dimensional adaptive threshold $\boldsymbol{\tau} = \big[\tau_k\big]_{k=1}^{K_s}$, we assemble a refined dataset $\mathbb{D}_{\tau}$ of selected target samples, as shown below:


\vspace{-2.5mm}
\begin{equation}
\label{high-conf-data}
    \mathbb{D}_{\tau} \leftarrow
   \big\{\big( \textbf{\textit{x}}^j_t,\hat{\boldsymbol{p}}^j_t\big) \hspace{1mm}\big| \hspace{1mm} \big( \textbf{\textit{x}}^j_t,\hat{\boldsymbol{p}}^j_t\big) \in \mathbb{D}_t, \hspace{1mm}\max(\hat{\boldsymbol{p}}^j_t) \geq \tau_k \big\}_{j = 1}^{n_t}
\end{equation}
\vspace{-8mm}

\begin{equation}
\begin{gathered}
 \label{threshold}    
    \tau_k \leftarrow \min\bigg(e^{\big(\frac{\tilde{p}_{t,k}}{\tilde{p}_{s,k}}\big)^\zeta}-1,1\bigg) \cdot \tilde{p}_{s,k}, \hspace{2mm} \forall k \in \{1, \cdots, K_s\}\\
    \tilde{p}_{o,k} \leftarrow \frac{\sum_{i=1}^{n_{o}} 
    \underset{[argmax \text{ }\hat{\boldsymbol{p}}^j_{o} = k]}{\mathbbm{1}} max(\hat{\boldsymbol{p}}^j_{o})}{\sum_{i=1}^{n_{o}} \underset{[argmax \text{ }\hat{\boldsymbol{p}}^j_{o} = k]}{\mathbbm{1}}}, \hspace{2mm} o \in \{s,t\} 
\end{gathered}   
\end{equation}

\vspace{2mm}
\noindent
The symbol $\tilde{p}_{o,k}$ denotes the average confidence $\mathcal{C}_p$ assigns to its predictions for the $k^{th}$ category ($l_k$) in domain $o \in$ \{source, target\}. Initially, $\tilde{p}_{t,k}$ is typically lower than $\hat{p}_{s,k}$. If $\tau_k$ relies solely on $\hat{p}_{s,k}$, the count of target samples in $\mathbb{D}_{\tau}$ might dwindle, especially at initial training stages, compromising target supervision performance. To mitigate this, we adjust $\tau_k$ with a non-linear function (first term on the R.H.S.) influenced by the user-set $\zeta$ regulator, lowering its value if the target's confidence falls below the source's. When $\tilde{p}_{t,k} \geq \tilde{p}_{s,k}$, $\tau_k$ equals the source's average confidence for the $l_k$ class.

\vspace{2mm}
\subsubsection{\textbf{Maximizing Inter-Class Separation}}
\label{inter}

A compact clustering of samples in the latent space, based on category-level distributions, is essential for improved classification. This involves ensuring \textit{different class labels occupy distinct distributions while similar labels cluster within their distributions, irrespective of the domains}. This objective is partially realized with $\mathscr{L}_{inter}$ (see eq. \ref{inter}), which seeks to separate two distinct class-conditional distributions by maximizing the $L_2$ distance between their class-wise mean latent embeddings of samples, across domains (eq. \ref{l2-distance}). Additionally, it maximizes the \textit{average Hausdorff distance} using an $L_2$-norm (eq. \ref{hausdorff}) between samples from two distinct classes to capture the geometric relations between distributions, enhancing category-level separation. $\mathscr{L}_{inter}$ operates on datasets $\mathbb{D}_s$ and $\mathbb{D}_{\tau}$ by partitioning them into categories in the label-index set $\mathcal{L}_{\mathbb{D}_{\tau}}$ of size $K_{\tau}$ $\big(\hspace{0.25mm} \mathcal{L}_{\mathbb{D}_{\tau}} = \{argmax\text{ }\hat{\boldsymbol{p}}^j_{t} \hspace{1mm} | \hspace{1mm} (\textbf{\textit{x}}^j_{t},\hat{\boldsymbol{p}}^j_{t}) \in \mathbb{D}_{\tau}\}_{j=1}^{n_t} \hspace{0.25mm}\big)$. 
For each category $k \in \mathcal{L}_{\mathbb{D}_{\tau}}$, the refined source and target datasets $(\mathbb{D}_{s_k}$ and $ \mathbb{D}_{\tau_k}$ respectively, in eq. \ref{inter}$)$, are formed using samples from class $l_k \in \mathcal{Y}_s$. Hyper-parameters $\alpha$ and $\beta$ balance the contribution of cross-domain and within-domain terms.

\vspace {-4mm}   
    \begin{multline}
    \label{inter}
        \mathscr{L}_{inter}(\theta_{\mathcal{E}}) = \\ 
        \frac{\alpha}{K_{\tau}(K_{\tau}-1)} \sum_{k \in \mathcal{L}_{\mathbb{D}_{\tau}}}\sum_{\substack{k' \in \mathcal{L}_{\mathbb{D}_{\tau}}\\k \neq k'}} \!\!\! \Big[d_e(\mathbb{D}_{\tau_k},\mathbb{D}_{\tau_{k'}})+ d_h(\mathbb{D}_{\tau_k},\mathbb{D}_{\tau_{k'}})\Big] \hspace{2mm}+\\
        \frac{\beta}{K_{\tau}(K_{\tau}-1)} \sum_{\substack{k \in \mathcal{L}_{\mathbb{D}_{\tau}}}}\sum_{\substack{k' \in \mathcal{L}_{\mathbb{D}_{\tau}}\\k \neq k'}} \!\!\! \Big[d_e(\mathbb{D}_{s_k},\mathbb{D}_{\tau_{k'}})+ d_h(\mathbb{D}_{s_k},\mathbb{D}_{\tau_{k'}})\Big]
    \end{multline}

\vspace {-5mm}   
    \begin{multline}
    \label{l2-distance}
    d_e(\mathbb{D},\mathbb{D}') = \bigg|\bigg|\frac{1}{|\mathbb{D}|}\sum_{\textbf{\textit{x}} \in \mathbb{D}}\mathcal{E}(\textbf{\textit{x}}) - \frac{1}{|\mathbb{D}'|}\sum_{\textbf{\textit{x}}' \in \mathcal{\mathbb{D}}'}\mathcal{E}(\textbf{\textit{x}}')\bigg|\bigg|_2
    \end{multline}

\vspace {-6mm}
    \begin{multline}
    \label{hausdorff}
    d_h(\mathbb{D},\mathbb{D}') = \frac{1}{2}\bigg[\frac{1}{|\mathbb{D}|}\sum_{\textbf{\textit{x}} \in \mathbb{D}} \min\limits_{\substack{\textbf{\textit{x}}' \in \mathbb{D}'}}\big|\big|\mathcal{E}(\textbf{\textit{x}})-\mathcal{E}(\textbf{\textit{x}}')\big|\big|_2 \hspace{2mm}+\\
    \frac{1}{|\mathbb{D}'|}\sum_{\textbf{\textit{x}}' \in \mathbb{D}'} \min\limits_{\substack{\textbf{\textit{x}} \in \mathbb{D}}}\big|\big|\mathcal{E}(\textbf{\textit{x}}')-\mathcal{E}(\textbf{\textit{x}})\big|\big|_2
    \bigg]    
    \end{multline}

\vspace{2mm}
\subsubsection{\textbf{Maximizing Intra-Class Compactness}}
\label{intra}

Previously, we underscored the importance of enhancing class distinction to prevent misclassification by classifiers. This section introduces the intra-class objective, $\mathscr{L}_{intra}$, that aims to group together samples from the same class, ensuring tight clusters. The goal is to reduce the distance between the latent representations of samples in the same class $l_k$, $k \in \mathcal{L}_{\mathbb{D}_{\tau}}$, without considering their originating domain. The objective $\mathscr{L}_{intra}$ acts on an aggregated dataset, represented as $\mathbb{D} = \bigcup_{k \in \mathcal{L}_{\mathbb{D}_{\tau}}} \mathbb{D}_k$, with each $\mathbb{D}_{k} = \mathbb{D}_{s_k} \cup \mathbb{D}_{\tau_k}$. The detailed objective is outlined below:

\vspace{-0.4cm}    
    \begin{multline}
    \label{intra}
        \mathscr{L}_{intra}(\theta_{\mathcal{E}}) = \\
        \frac{1}{K_s} \sum_{k = 1}^{K_s} \! \Bigg[ \frac{1}{{|\mathbb{D}_k|}({|\mathbb{D}_k|-1})} \! \sum_{\textbf{\textit{x}}^i \in \mathbb{D}_k} \sum_{\substack{\textbf{\textit{x}}^j \in \mathbb{D}_k\\j \neq i}} \!\! \Big|\Big| \mathcal{E}(\textbf{\textit{x}}^i) - \mathcal{E}(\textbf{\textit{x}}^j)\Big|\Big|_2\Bigg]
    \end{multline}

\subsubsection{\textbf{Entropy Minimization of Target Samples}}

The initial stages of a classification process with pre-existing domain shifts witness significant negative effects, such as a decrease in the classifier's certainty due to noisy pseudo-target labels. As a result, the classifier's predictions tend to produce low and uniform probabilities across all classes, including the ground-truth class of the sample. To mitigate this issue, we utilize the principle of entropy minimization on the target samples in $\mathbb{D}_t$. This objective is formulated as:

\vspace{-0.1cm}
\begin{equation}
\label{ent}
    \mathscr{L}_{ent}(\theta_{\mathcal{C}},\theta_{\mathcal{E}}) = 
    - \frac{1}{n_t} \sum\limits_{j=1}^{n_t} \sum\limits_{k=1}^{K_s} {\boldsymbol{p}_t^j}_k \hspace{1mm} log\big({\boldsymbol{p}_t^j}_k\big)
\end{equation}

\noindent
The classifier output ${\boldsymbol{p}_t^j}_k$ in eq. \ref{ent} refers to the predicted probability of sample $\textbf{\textit{x}}^j_t$ belonging to class $l_k$.

\vspace{2mm}
\subsubsection{\textbf{Overall Objective}}

To summarize, the overall objective is represented as follows:

\vspace{-3mm}
\begin{multline}
\label{overall_obj}
\min\limits_{(\theta_{\mathcal{C}},\theta_{\mathcal{E}})} \Big\{
\mathscr{L}_{ce}(\theta_{\mathcal{C}},\theta_{\mathcal{E}}) + \eta \mathscr{L}_{comp}(\theta_{\mathcal{C}},\theta_{\mathcal{E}}) -  \mathscr{L}_{inter}({\theta_{\mathcal{E}}}) + \\
\delta \mathscr{L}_{intra}({\theta_{\mathcal{E}}}) + \mathscr{L}_{ent}(\theta_{\mathcal{C}},\theta_{\mathcal{E}})\Big\}
\end{multline}

\noindent
Here, $\eta$ and $\delta$ are user-defined hyperparameters regulating the contribution of each objective in the learning process.

\vspace{0.3mm}
\section{\textbf{Experiments}}
\label{experiments}

\noindent
In this section, we detail an exhaustive evaluation of our proposed model in comparison to existing state-of-the-art methods, utilizing two benchmark datasets. Our evaluation spans various \textit{pda} scenarios, incorporating several adaptation tasks for an in-depth review. Consistent with established evaluation standards \cite{cao2018partial2,tzeng2017adversarial,cao2018partial}, we employ classification accuracy as the key metric, incorporating all labeled source data and unlabeled target data for the adaptation tasks. We also offer an in-depth analysis of model performance, shedding light on the impact of the \textit{Complement Entropy Objective}, \textit{Intra/Inter-Class Distribution Optimization}, and the \textit{Robust Pseudo-label-based Target Supervision} components. Subsequent sections present the outcomes of our experiments and an ablation analysis of the aforementioned modules.


\vspace{1mm}
\begin{table}[htbp!]
\centering
\begin{tabular}{l | c c c c c c}
  \hline
 \textbf{Dataset} & $ \hspace{2mm} \mathbf{\gamma} \hspace{2mm}$ & $\hspace{2mm} \mathbf{\eta} \hspace{2mm}$ & $\hspace{2mm} \mathbf{\alpha} \hspace{2mm}$ & $\hspace{2mm} \mathbf{\beta} \hspace{2mm}$ & $\hspace{2mm} \mathbf{\delta} \hspace{2mm}$ & $\hspace{2mm} \mathbf{\zeta} \hspace{2mm}$\\
  \hline
 \textbf{\textit{Office-31}} & 0.7 & 6 & 0.4 & 1 & 1.5 & 3\\
 \textbf{\textit{Office-home}} & 0.3 & 2 & 0.4 & 1 & 1.5 & 3\\
  \hline
\end{tabular}
\vspace{2mm}
\caption{Parameter settings for model evaluation.}
\label{parameter-settings}
\vspace{-7mm}
\end{table}

\begin{table*}[htbp!]
\centering
\begin{adjustbox}{width=0.55\textwidth}
\begin{tabular}{c | c c c c c c | c} 
\hline
\textbf{Method} & \textbf{A $\rightarrow$ D} & \textbf{A $\rightarrow$ W} & \textbf{D $\rightarrow$ A} & \textbf{D $\rightarrow$ W} & \textbf{W $\rightarrow$ A} & \textbf{W $\rightarrow$ D} & \textbf{Avg.}\\
\hline
\textbf{Resnet-50}\cite{he2016deep} & 83.44 & 75.59 & 83.92 & 96.27 & 84.97 & 98.09 & 87.05\\
\textbf{DANN}\cite{ganin2016domain} & 81.53 & 73.56 & 82.78 & 96.27 & 86.12 & 98.73 & 86.50\\
\textbf{ADDA}\cite{tzeng2017adversarial} & 83.41 & 75.67 & 83.62 & 95.38 & 84.25 & 99.85 & 87.03\\
\textbf{PADA}\cite{cao2018partial2} & 82.17 & 86.54 & 92.69 & 99.32 & 95.41 & {\color{red}\textbf{100.00}} & 92.69\\
\textbf{IWAN}\cite{zhang2018importance} & 90.45 & 89.15 & 95.62 & 99.32 & 94.26 & 99.36 & 94.69\\
\textbf{SAN}\cite{cao2018partial} & 94.27 & 93.90 & 94.15 & 99.32 & 88.73 & 99.36 & 94.96\\
\textbf{ETN}\cite{cao2019learning} & 95.03 & 94.52 & {\color{red}\textbf{96.21}} & {\color{red}\textbf{100.00}} & 94.64 & {\color{red}\textbf{100.00}} & 96.73\\
\hline
\textbf{Proposed Model} & {\color{red}\textbf{97.13}} & {\color{red}\textbf{97.58}} & 95.93 & {\color{red}\textbf{100.00}} & {\color{red}\textbf{95.82}} & {\color{red}\textbf{100.00}} & {\color{red}\textbf{97.74}}\\
\hline
\end{tabular}
\end{adjustbox}
\vspace{2mm}
\caption{Accuracy of classification (\%) achieved for partial domain adaptation tasks on the \textit{Office-31} dataset (Resnet-50 backbone)}
\label{office-31}
\vspace{-4mm}
\end{table*}

\begin{table*}[htbp!]
\centering
\begin{adjustbox}{width=1\textwidth}
\begin{tabular}{c | c c c c c c c c c c c c | c}
 \hline
\textbf{Method} & \textbf{Ar $\rightarrow$ Cl} & \textbf{Ar $\rightarrow$ Pr} & \textbf{Ar $\rightarrow$ Rw} & \textbf{Cl $\rightarrow$ Ar} & \textbf{Cl $\rightarrow$ Pr} & \textbf{Cl $\rightarrow$ Rw} & \textbf{Pr $\rightarrow$ Ar} & \textbf{Pr $\rightarrow$ Cl} & \textbf{Pr $\rightarrow$ Rw} & \textbf{Rw $\rightarrow$ Ar} & \textbf{Rw $\rightarrow$ Cl} & \textbf{Rw $\rightarrow$ Pr} & \textbf{Avg.}\\ 
 \hline
\textbf{Resnet-50}\cite{he2016deep} & 46.33 & 67.51 & 75.87 & 59.14 & 59.94 & 62.73 & 58.22 & 41.79 & 74.88 & 67.40 & 48.18 & 74.17 & 61.35\\
\textbf{DANN}\cite{ganin2016domain} & 43.76 & 67.90 & 77.47 & 63.73 & 58.99 & 67.59 & 56.84 & 37.07 & 76.37 & 69.15 & 44.30 & 77.48 & 61.72\\
\textbf{ADDA}\cite{tzeng2017adversarial} & 45.23 & 68.79 & 79.21 & 64.56 & 60.01 & 68.29 & 57.56 & 38.89 & 77.45 & 70.28 & 45.23 & 78.32 & 62.82\\
\textbf{PADA}\cite{cao2018partial2} & 51.95 & 67.00 & 78.74 & 52.16 & 53.78 & 59.03 & 52.61 & 43.22 & 78.79 & 73.73 & 56.60 & 77.09 & 62.06\\
\textbf{DRCN}\cite{li2020deep} & 54.00 & 76.40 & 83.00 & 62.10 & 64.50 & 71.00 & 70.80 & 49.80 & 80.50 & 77.50 & 59.10 & 79.90 & 69.00\\
\textbf{IWAN}\cite{zhang2018importance} & 53.94 & 54.45 & 78.12 & 61.31 & 47.95 & 63.32 & 54.17 & 52.02 & 81.28 & 76.46 & 56.75 & 82.90 & 63.56\\
\textbf{SAN}\cite{cao2018partial} & 44.42 & 68.68 & 74.60 & 67.49 & 64.99 & {\color{red}\textbf{77.80}} & 59.78 & 44.72 & 80.07 & 72.18 & 50.21 & 78.66 & 65.30\\
\textbf{ETN}\cite{cao2019learning} & 59.24 & 77.03 & 79.54 & 62.92 & 65.73 & 75.01 & 68.29 & 55.37 & 84.37 & 75.72 & 57.66 & {\color{red}\textbf{84.54}} & 70.45\\

 \hline
\textbf{Proposed Model} & {\color{red}\textbf{61.54}} & {\color{red}\textbf{83.45}} & {\color{red}\textbf{89.12}} & {\color{red}\textbf{70.24}} & {\color{red}\textbf{74.46}} & 77.62 & {\color{red}\textbf{70.82}} & {\color{red}\textbf{55.66}} & {\color{red}\textbf{85.70}} & {\color{red}\textbf{78.16}} & {\color{red}\textbf{59.44}} & 83.23 & {\color{red}\textbf{74.12}} \\
 \hline
\end{tabular}
\end{adjustbox}
\vspace{1mm}
\caption{Accuracy of classification (\%) achieved for partial domain adaptation tasks on the \textit{Office-home} dataset (Resnet-50 backbone)}
\label{Office-home}
\vspace{-8mm}
\end{table*}

\subsection{\textbf{Datasets}} 
\label{datasets}

\noindent
To evaluate the target classification performance in a cross-domain setup, we employ two commonly used image datasets for domain adaptation: \textit{Office-home} \cite{venkateswara2017deep} and \textit{Office-31} \cite{saenko2010adapting}.\\

\noindent
\textbf{Office-31:} The \textit{Office-31} dataset \cite{saenko2010adapting} comprises RGB images from three distinct domains: Amazon (A), DSLR (D), and Webcam (W). These images are classified into 31 categories. To establish a partial domain adaptation setup, we adopt the standard protocol proposed by Cao et al. \cite{cao2018partial2}, where the target dataset includes samples from 10 categories. To conduct a thorough evaluation, we test the proposed model across multiple adaptation tasks on the following source-target pairs: A$\rightarrow$D, A$\rightarrow$W, D$\rightarrow$A, D$\rightarrow$W, W$\rightarrow$A, and W$\rightarrow$D.\\

\vspace{-2mm}
\noindent
\textbf{Office-home:} \textit{Office-home} \cite{venkateswara2017deep} is a larger dataset that comprises RGB images from four domains, namely Artistic (Ar), Clip Art (Cl), Product (Pr), and Real-world (Rw). In line with the evaluation setup presented for \textit{Office-31}, we follow the same protocol and create the source and target datasets with 65 and 25 categories, respectively. To conduct a thorough evaluation, we consider 12 different adaptation tasks, namely Ar$\rightarrow$Cl, Ar$\rightarrow$Pr, Ar$\rightarrow$Rw, Cl$\rightarrow$Ar, Cl$\rightarrow$Pr, Cl$\rightarrow$Rw, Pr$\rightarrow$Ar, Pr$\rightarrow$Cl, Pr$\rightarrow$Rw, Rw$\rightarrow$Ar, Rw$\rightarrow$Cl, and Rw$\rightarrow$Pr.

\subsection{\textbf{Implementation}}
\label{implementation}

\noindent
We conducted our experiment on an Nvidia 3090-Ti GPU with 24 GB memory, utilizing PyTorch. We employed a Resnet-50, pre-trained on Imagenet, as the primary model backbone, which was then fine-tuned with source samples. Built atop this backbone, the feature encoder $ \mathcal{E}(\hspace{0.5mm}\cdot\hspace{0.5mm}) $ omits the last dense layer and incorporates two fully-connected layers, with a hidden-layer size of 1024, followed by ReLU activations, with 0.1 dropout probability. This encoder output layer yields 512-dimensional latent representations, further processed by the neural network $ \mathcal{C}(\hspace{0.5mm}\cdot\hspace{0.5mm}) $ and the prototype classifier $ \mathcal{C}_p(\cdot, \cdot) $. $ \mathcal{C}(\hspace{0.5mm}\cdot\hspace{0.5mm}) $ is a two-layer dense neural network with hidden layer output dimensions of 512. The output dimensions vary per dataset: 31 for \textit{Office-31} and 65 for \textit{Office-home}.

The model is trained for 950 epochs using the ADAM optimizer with a learning rate of $1e-4$. Parameters $ \gamma $ and $ \eta $, linked to the \textit{adaptive complement entropy objective}, were optimized for tasks A $ \rightarrow $ W and Ar $ \rightarrow $ Rw. The $ \omega $ parameter, affecting the centroid update in equation \ref{mean-calc}, is 0.1. The parameters $ \alpha $, $ \beta $, and $ \delta $, geared towards achieving intra-class compactness and inter-class separation, are fine-tuned on tasks A $ \rightarrow $ W using the \textit{Office-Home} dataset and are maintained uniformly across all datasets. $ \zeta $, guiding the change of the $ \tau_k $ threshold as average target confidence nears the average source confidence for class $ k $, is set at 3. The fine-tuned parameter values utilized in the model are reported in table \ref{parameter-settings}. Target classification outputs from $\mathcal{C}(\hspace{0.5mm}\cdot\hspace{0.5mm})$ are reported for model evaluation.

\vspace{-4mm}
\begin{figure}[htbp!]
    \centering
    \includegraphics[width = 0.98\linewidth]{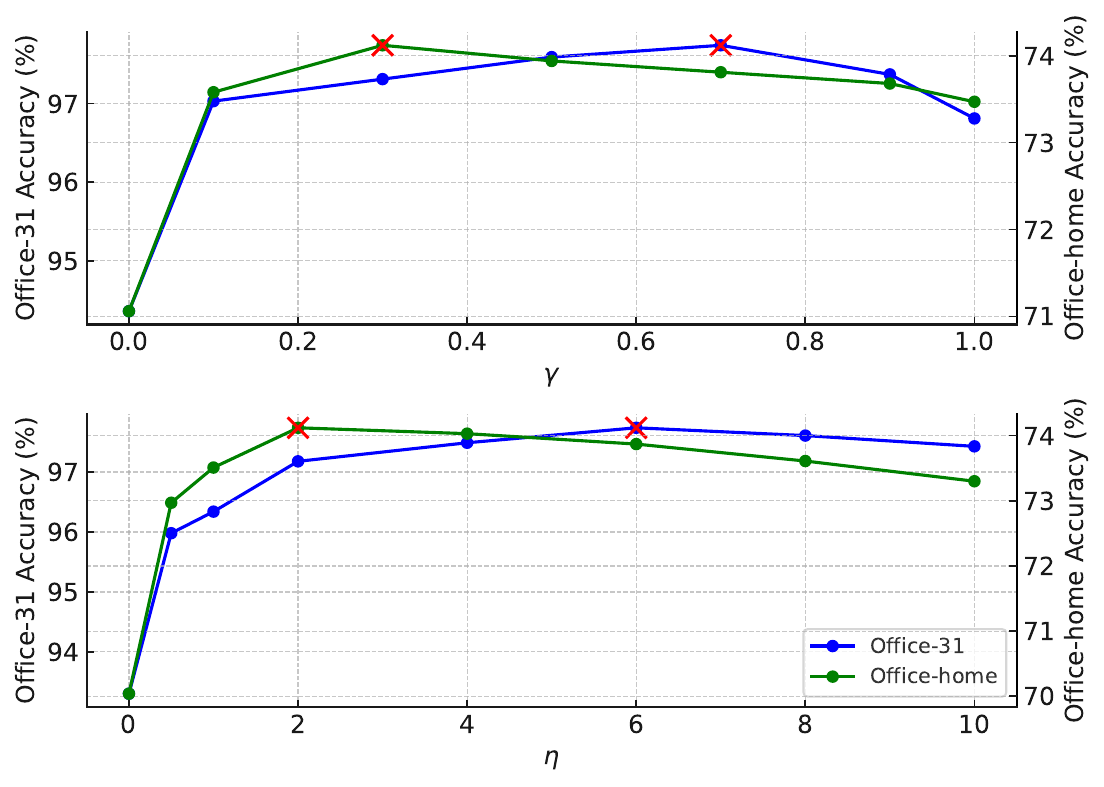}
    \vspace{-4.5mm}
    \caption{Average accuracy \% for $\gamma$ and $\eta$ values on \textit{Office-31} and \textit{Office-home}.}    
    \label{ablation_pic}  
    \vspace{-5mm}
\end{figure}

\vspace{2mm}
\subsection{\textbf{Comparison Models}}
\label{comparison_models}

\noindent
To evaluate our proposed method against state-of-the-art models for both closed-set and partial-domain adaptation tasks, we use the target classification accuracy metric and utilize all samples from both the source and target datasets ($\mathbb{D}_s$ and $\mathbb{D}_t$). The models we compare against include Domain Adversarial Neural Network (DANN) \cite{ganin2016domain}, Partial Adversarial Domain Adaptation (PADA) \cite{cao2018partial2}, Adversarial Discriminative Domain Adaptation (ADDA) network \cite{tzeng2017adversarial}, Importance Weighted Adversarial Nets (IWAN) \cite{zhang2018importance}, Example Transfer Network (ETN) \cite{cao2019learning}, Selective Adversarial Network (SAN) \cite{cao2018partial}, and Deep Residual Correction Network (DRCN) \cite{li2020deep}. To emphasize the negative transfer issue in the DANN model (which is exclusively capable of solving closed-set adaptation tasks), we include the classification accuracy of Resnet-50 \cite{he2016deep} trained exclusively on the target data in a supervised manner.

\begin{figure*}[htbp!]
    \centering
    \includegraphics[width = 1\linewidth]{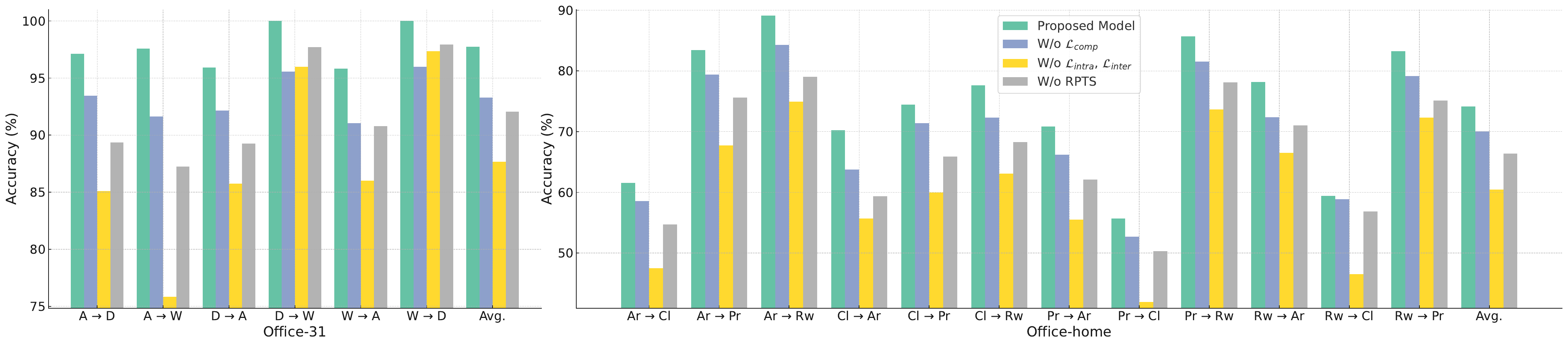}
    \vspace{-8mm}
    \caption{Reported accuracies over cross-domain tasks after suppressing individual components, illustrating their respective impacts to the overall performance.}    
    \label{ablation_combined}  
    \vspace{-5mm}
\end{figure*}

\vspace{0.1cm}
\subsection{\textbf{Classification Results}}
\label{classification_results}

\noindent
The target classification accuracies for the \textit{Office-31} and \textit{Office-home} benchmark datasets are presented in tables \ref{office-31} and \ref{Office-home}, respectively. It is noteworthy that the accuracy values for Resnet-50 \cite{he2016deep} and DANN \cite{ganin2016domain} in tasks A $\rightarrow$ W, A $\rightarrow$ D, D $\rightarrow$ A (table \ref{office-31}) and Ar $\rightarrow$ Cl, Cl $\rightarrow$ Pr, Pr $\rightarrow$ Ar, Pr $\rightarrow$ Cl, and Rw $\rightarrow$ Cl (table \ref{Office-home}) indicate the existence of the negative transfer problem; the standard DANN model, designed for addressing closed-set domain adaptation problems, fails to filter out the impact of samples from classes exclusive to the source domain ($\mathcal{Y}_s \setminus \mathcal{Y}_t$), thereby impeding its ability to achieve improved accuracy. Conversely, our proposed model, tailored exclusively for the \textit{pda} task, seeks to reduce negative transfer. It does so by curating a structured ``latent space" that distinctly isolates private class details from shared class data.

Our approach differs from other methods \cite{tzeng2017adversarial,cao2018partial2, cao2018partial,cao2019learning} that rely exclusively on the class/sample importance weight estimation from the outset of training. While many methods primarily focus on mitigating domain discrepancy, we aim to align the domain distributions without compromising feature distinctiveness. Empirical results presented in tables \ref{office-31} and \ref{Office-home} demonstrate the superiority of our proposed model, which achieves the highest classification accuracies in 5 out of 6 tasks and 10 out of 12 tasks, respectively, while also yielding the highest average accuracy across both datasets.

\vspace{0.1mm}
\subsection{\textbf{Parameter Sensitivity}}
\label{parameter_sensitivity}

\noindent
The trade-off parameters $\gamma$ and $\eta$, controlling the complement entropy objective (eq, \ref{comp}, \ref{overall_obj}), play a critical role in the model's learning process. While $\gamma$ controls the emphasis placed on samples based on classification confidence, giving priority to uncertain but confident samples that yield smaller cross-entropy loss, $\eta$ regulates the contribution of $\mathscr{L}_{comp}$ to the overall objective. In figure \ref{ablation_pic}, we report the mean accuracy of the proposed classification network for various values of $\gamma$ and $\eta$ on the \textit{Office-31} and \textit{Office-home} datasets. Our observations indicate that accuracy values remain within an acceptable range ($\leq 1.8\%$) for $\gamma$ and $\eta$ values $>0$, indicating that the approach is less sensitive to variations in these parameters.

\subsection{\textbf{Ablation Analysis}}
\label{ablation_analysis}

\noindent
In this section, we performed an ablation study by disabling each component and assessing the subsequent performance. This helps gauge the significance of each element in our proposed network. The analysis specifics are outlined below.
\vspace{1mm}

\begin{itemize}
\item \textbf{W/o $\mathscr{L}_{comp}$}: While objectives $\mathscr{L}_{inter}$ and $\mathscr{L}_{intra}$ enhance inter-class separability and intra-class cohesion w.r.t source-target and target-target interactions, they don't explicitly manage source sample interactions to avoid computational overhead. The $\mathscr{L}_{comp}$ objective aims to fill this gap by efficiently creating distinct source clusters. This is achieved by ensuring a uniform distribution of low-prediction probabilities among complement classes, making it difficult for an incorrect class to challenge the ground-truth class. We suppressed $\mathscr{L}_{comp}$ from the overall loss objective by setting $\eta$ to 0 to test this hypothesis. As shown in fig. \ref{ablation_combined}, the average classification accuracy drops significantly ($>4$\%), which confirms the effectiveness of this module.
\vspace{1mm}

\item \textbf{W/o $\mathscr{L}_{intra}, \mathscr{L}_{inter}$}: 
We posited that achieving alignment of class-conditional distributions is as crucial as reducing the domain shift between the source and target domains. To this end, we proposed the $\mathscr{L}_{inter}$ objective, which maximizes inter-category distance in the latent space by exploring beyond the first-order moments of the distributions. Additionally, we employed the $\mathscr{L}_{intra}$ objective to enhance intra-class compactness across domains. To evaluate their impact, we set $\alpha$, $\beta$, and $\delta$ to 0. The accuracy results in fig. \ref{ablation_combined} show a significant drop in average classification accuracy, with drops of over $10\%$ and $18\%$ for the \textit{Office-31} and \textit{Office-home} datasets, respectively.
\vspace{2mm}

\item \textbf{W/o} \textit{RPTS}: 
We incorporate a pseudo-labeling technique named ``\textit{Robust Pseudo-label-Based Target Supervision (RPTS)}" in our method. In the early stages of model training, many generated pseudo-labels might be noisy, potentially hindering the learning process. To counter this, we select a subset of target samples with prediction probabilities exceeding an adaptive threshold for supervision. This threshold is set based on the average confidence of classifier predictions for both target and source samples. To assess the impact of our \textit{RPTS} module, we bypassed this technique and, instead, conducted model supervision using all target pseudo-labels produced by the neural network classifier, $ \mathcal{C}(\hspace{0.5mm}\cdot\hspace{0.5mm}) $. This meant replacing dataset $ \mathbb{D}_{\tau} $ with $ \mathbb{D}_t $ in equations \ref{class} and \ref{comp} for $ \mathscr{L}_{ce} $ and $ \mathscr{L}_{comp} $, respectively. The observed decline in accuracy rates ($\sim$ 5.4\% for \textit{Office-31} and $\sim$ 10.8\% for \textit{Office-home} as shown in fig. \ref{ablation_combined}) underscores the effectiveness of the \textit{RPTS} module in target supervision.

\end{itemize}

\vspace{1mm}
\noindent
The results indicate that the objectives that aim to optimize class distribution (inter-class separation and intra-class compactness) have the greatest impact on performance, followed by the \textit{RPTS} module. The complement entropy objective contributes significantly, as its removal resulted in notable performance drops in all tasks across both datasets.

\vspace{1mm}
\section{\textbf{Conclusion}}

\noindent
This work presents a simple yet effective classification approach tailored for partial domain adaptation tasks. Instead of relying on the existing class/sample re-weighting-based techniques, our strategy underscores the significance of category-level feature alignment. We employ objectives that aim to obtain distinct category-level distributions by exploring beyond first-order moments and optimizing within-class compactness while aligning domain distributions. The complement entropy objective reduces classification ambiguity, producing well-separated category distributions. Furthermore, a robust pseudo-labeling method is proposed with an adaptive threshold to select target samples based on prediction confidence for effective target supervision. Testing on two benchmarks against state-of-the-art models and subsequent ablation analysis confirms our approach's superiority in all benchmark tasks.


\printbibliography

\end{document}